Data and text mining

# Assigning function to protein-protein interactions: a weakly supervised BioBERT based approach using PubMed abstracts

Aparna Elangovan[1], Melissa Davis[2,3], and Karin Verspoor[1,*]

[1]School of Computing and Information Systems, The University of Melbourne, Melbourne, Australia
[2]The Walter and Eliza Hall Institute of Medical Research, Melbourne, Australia
[3]Department of Clinical Pathology, Faculty of Medicine, Dentistry & Health Sciences, The University of Melbourne, Australia

*To whom correspondence should be addressed.

## Abstract

**Motivation:** Protein-protein interactions (PPI) are critical to the function of proteins in both normal and diseased cells, and many critical protein functions are mediated by interactions. Knowledge of the nature of these interactions is important for the construction of networks to analyse biological data. However, only a small percentage of PPIs captured in protein interaction databases have annotations of function available, e.g. only 4% of PPI are functionally annotated in the IntAct database. Here, we aim to label the function type of PPIs by extracting relationships described in PubMed abstracts.

**Method:** We create a weakly supervised dataset from the IntAct PPI database containing interacting protein pairs with annotated function and associated abstracts from the PubMed database. We apply a state-of-the-art deep learning technique for biomedical natural language processing tasks, BioBERT, to build a model – dubbed PPI-BioBERT – for identifying the function of PPIs. In order to extract high quality PPI functions at large scale, we use an ensemble of PPI-BioBERT models to improve uncertainty estimation and apply an interaction type-specific threshold to counteract the effects of variations in the number of training samples per interaction type.

**Results:** We scan 18 million PubMed abstracts to automatically identify 3253 new typed PPIs, including phosphorylation and acetylation interactions, with an overall precision of ~46% (87% for acetylation) based on a human-reviewed sample. This work demonstrates that analysis of biomedical abstracts for PPI function extraction is a feasible approach to substantially increasing the number of interactions annotated with function captured in online databases.

**Contact:** karin.verspoor@unimelb.edu.au

**Supplementary information:** Supplementary data are available at *Bioinformatics* online.

**Availability:** We publish our source code and dataset on https://github.com/elangovana/PPI-typed-relation-extractor.

## 1 Introduction

Critical biological processes, such as signaling cascades and metabolism, are regulated by protein interactions that modify proteins in order to modulate their stability or activity. Protein-protein interactions (PPIs) are collected in large online repositories such as IntAct (Orchard et al. 2013) and HPRD (Mishra et al. 2006). These databases use manual curation to collect interactions from the literature, and contain many thousands of interactions, and these networks are frequently used to interpret the impact of high-throughput molecular data (Genovesi et al. 2013). However, most PPIs are not annotated with a function, which makes it difficult to identify the consequences of network disruptions observed in disease. For example, we found the IntAct database has >100,000 human PPIs, but less than 4% (3381) of these are annotated with functions such as phosphorylation, acetylation or methylation. Understanding the function of a PPI is critical for researchers to determine the impact of network perturbations and downstream biological consequences.

Many proteins fulfill their function by interacting with other proteins through post-translational modifications (PTM) or by forming protein complexes. Post-translational modification of proteins refers to the chemical changes that proteins may undergo after translation. Text mining has been used to curate PTMs (Veuthey et al. 2013), including in BioNLP community shared tasks that benchmark the effectiveness of current text mining methods. However, these tasks have focused on curating details of individual proteins, such as annotating the type of modification (e.g. phosphorylation), the modified amino acid and its position in the protein sequence, and the identity of the protein that is modified (Verspoor et al. 2012, Veuthey 2013). These tasks do not address identification of the type of interaction between a *pair* of proteins, and therefore do not assist in providing detailed edge type information.

Existing text mining approaches for the extraction of PPIs typically do not aim to capture the function of interactions. This is because most are based on the annotated PPI datasets detailed in Pyysalo et al. (2008), which focus on identifying pairs of interacting proteins, but not the nature of the interaction between the pair. Here we extend this work, introducing a text mining approach for extraction of protein pairs along with their functions. We focus on extracting seven (7) PTM functions (phosphorylation, dephosphorylation, methylation, demethylation, ubiquitination, deubiquitination, and acetylation), through analysis of journal abstracts. We approach this through application of state-of-the-art deep learning methods to a new dataset of PubMed abstracts with annotations of a) two interacting proteins along with b) the type of interaction between a pair, constructed using a distant supervision (Craven et al. 1999) methodology.

### 1.1 PPI relation extraction through text mining

High quality automatic extraction of PPIs from text depends on a) availability of sufficient volumes of training data, and b) advancements in machine learning techniques. Here, we briefly review these two aspects in the context of our objective to extract typed PPI from biomedical journals. Evaluation of text mining methods for PPI tasks generally has made use of the AIMed (Bunescu et al. 2005) or BioInfer (Pyysalo et al. 2007) corpora, which have sentence level gold standard annotations of interacting protein pairs. The AIMed (Bunescu 2005) dataset consists of 225 Medline abstracts, of which 200 describe interactions between human proteins and 25 do not refer to any interaction. AIMed does not have any PPI interaction function annotated, just that 2 proteins interact with one another. BioInfer (Pyysalo 2007) has a single interaction type, binding, annotated over the dataset.

Creating gold standard training data with sentence-level, fine-grained annotations, is a manual, labor-intensive task and is a limiting factor in applying machine learning to new domains or tasks. Weakly supervised datasets can make use of existing data sources or require annotations that are less labor intensive. Zhou (2017) describe 3 categories of weak supervision, incomplete, inexact and inaccurate supervision. In the case of incomplete supervision, only a subset of training data has labels while the other data remains unlabeled. Inexact supervision is the case where only coarse-grained labels are available. Inaccurate supervision occurs where the labels are not always ground truth. Being able to leverage one or more these types of weakly supervised datasets is key to using machine learning in new domains or for new tasks. We will describe below how we apply this approach to support the extraction of typed PPIs from text sources, overcoming the lack of gold-standard annotation data for training.

Apart from training data, the next aspect to consider for reliable PPI extraction from text is the effectiveness of the machine learning method employed to learn a text mining model. The broad categories of deep learning architectures that have been applied for extracting PPI include Convolutional Neural Network (CNN), some form of Recurrent Neural Network (RNN) such as Bidirectional long short term memory network (BiLSTM ) and attention networks (Hsieh et al. 2017, Peng et al. 2017, Yadav et al. 2018). Previously, deep learning methods for natural language processing (NLP) have encoded words using pre-trained word embeddings which are trained on methods such as Word2Vec (Mikolov et al. 2013). However, these representations do not capture context-specific meanings or polysemy (where a word can have two meanings). One of the techniques to take into account context is Bidirectional Encoder Representations from Transformers (BERT), which uses a multi-layer bidirectional transformer encoder (Devlin et al. 2019) and is pretrained on a masked language modelling task to learn contextual representations (contextual word embeddings). BioBERT (Lee et al. 2019) is an implementation of BERT based on biomedical text data, using the BERT architecture with weights initialized in pretraining before fine-tuning on biomedical corpora. We therefore build on BioBERT to create our PPI extraction model.

## 2 Methods

### 2.1 Constructing weakly-supervised labelled data

In this paper we use a weakly supervised approach, collecting PPIs annotated with functions and the PubMed identifiers associated to these interactions from the IntAct database (Orchard 2013). The PPIs we select from IntAct have functions covering 7 types of interactions: phosphorylation, dephosphorylation, methylation, demethylation, ubiquitination, deubiquitination, and acetylation. Our task is to identify the *type* of a known PPI interaction, rather than solely to determine *whether* two proteins interact.

PPI interactions are often mentioned across multiple sentences in a biomedical journal abstract, as opposed to a single sentence (Verga et al. 2018). In our experiment, we make use of *abstract-level* annotations, where a typed PPI interaction is associated with the whole abstract. In this approach, we assume the description of the interaction may appear anywhere in the abstract, including across multiple sentences. This coarse-grained annotation, derived from the structured associations between a typed PPI and a PubMed identifier specified in IntAct, falls into the *inexact* category of weak supervision (Zhou 2017) due to uncertainties around the precise location in the publication of statements supporting the database entry, as well as potential lack of completeness of the annotation using this strategy.

For the purposes of constructing our data set, we assume that the annotated PPI interaction is described in the abstract of the PubMed article given as the source of the interaction, although in practice it may occur elsewhere the text body of the article. We also assume that if a relationship between a pair of proteins has not been manually associated with a given PubMed identifier in the IntAct database, then that relationship is not expressed within that text; this enables us to specify negative data. However, it is important to note that this assumption will fail if not all interactions described in a given article abstract are curated. Hence our, data is noisy.

#### 2.1.1 Preprocessing

We obtain the dataset of PPI interactions for humans from IntAct database (Orchard 2013), a database that is part of the International Molecular Exchange (IMEX) (Orchard et al. 2012) Consortium. The database contains UniProt identifiers, protein aliases, interaction type (where available) for interactions, and the PubMed identifier of the paper describing the interaction.

Of 3381 PPI interactions that correspond to the seven main interaction types in IntAct (Orchard 2013), we filter out relationships that have more than two participants to retain 2,868 interactions (~80% of the interactions). We use GNormPlus (Wei et al. 2015) as the named entity recognition (NER) tool to recognize all protein entity mentions in the abstracts. We replace the gene/protein names in the abstract with the corresponding UniProt identifiers to normalize gene mentions. We then apply another filter to remove PPIs where the UniProt identifiers of both known participants do not exist in the normalized abstract. This is to minimize the noise in the weakly supervised dataset, based on the assumption that if the proteins are not explicitly mentioned in the abstract then it is highly likely (unless protein names are not recognized by the NER tool) that the abstract does not describe the relationship between those proteins. Finally, we split the dataset into train, test and validation sets such that they are stratified by interaction type and have unique PubMed ids in each set to avoid test set leakage.

The raw text input is a normalized abstract where the gene names are replaced by UNIPROT identifiers. In addition, we replace the UNIPROT identifiers of participant-1 and participant-2 with "PROTEIN1" and "PROTEIN2" during training and inference. Essentially "PROTEIN1" and "PROTEIN2" act as markers indicating the participants in the relationship.

#### 2.1.2 Negative sample generation

In our dataset, negative samples are protein pairs that are mentioned in the abstract but do not have a function referencing that paper annotated against the pair in IntAct. In order to generate negative samples, we identify protein mentions from the abstracts using GNormPlus (Wei 2015), which normalizes mentions to UniProt IDs, and consider all possible pairs of co-occurring proteins. If a given protein pair <p1, p2> is not annotated against <u>any</u> PPI relationship within the abstract in the IntAct database and the types annotated against the abstract are t1 and t2, then it is assumed that <p1, p2, t1> and <p1, p2, t2> form negative samples for that abstract.

It is important to emphasize that a negative sample does not mean that a given PPI relationship does not exist, but rather that *the abstract does not describe such a relationship*. It could also be a noisy negative sample, i.e. the abstract describes the functional relationship between pair, but it is simply not captured in the annotations. This implies that an untyped PPI interaction between p1 and p2 would form negative examples for all interaction types mentioned in the abstract.

We also create an untyped version of our dataset that ignores the type of interaction, in order to compare our approach with existing datasets and PPI extraction methods that do not consider type. In this dataset, if two proteins interact, regardless of the type of interaction, they are labelled as positive samples and all other samples form negative examples. The overall positive sample rate is approximately 33% for the train, test and validation set compared to around 17% positive sample rate in the AIMed dataset.

### 2.2 Deep learning-based PPI model learning

We fine-tune pretrained BioBERT v1.1 (Lee 2019) which is trained on a large collection of PubMed abstracts. We use this specific model rather than context-insensitive word embeddings such as (Chiu et al. 2016) used for prior deep learning methods for PPI extraction (Hsieh 2017, Peng 2017, Yadav 2018, Zhang et al. 2019) because it to takes into account appropriate contextualized word usages in biomedical corpora. We use PyTorch as the deep learning framework for training our models.

We apply fine-tuning to adapt BioBERT to the typed PPI extraction task, by adding a fully connected final layer to classify input texts into the 8 classes (7 PPI types + 1 negative class). We model this as a multi-class classification problem, and assume there is at most one type of PTM relationship between a protein pair. This results in a model we refer to as PPI-BioBERT. Fine-tuning starts with pretrained weights (rather than randomly initialized ones) and new task-specific weights are learned for the new task of typed PPI extraction. We chose to update the weights for all layers, as opposed to the last *n* layers) as part of fine-tuning BioBERT.

The original BioBERT is trained using a sentence as a training sample for the language modelling task, while we work with complete abstract texts. To utilize an entire abstract as a single training sample, we feed all the sentences within the abstract using including the full stop (end-of-sentence marker, the period) "." separating each sentence. This full stop is tokenized along with the sentence and fed into the BioBERT network.

### 2.3 Uncertainty estimation

In order to improve the probability estimate associated with each prediction, we use an ensemble of 10 PPI-BioBERT models (referred to as PPI-BioBERT x10) all trained with exactly the same hyperparameters and training data but capturing slightly different models due to Bernoulli (binary) dropout layers in the network. Bernoulli dropout (Srivastava et al. 2014) prevents overfitting by randomly dropping weights to zero with a probability p and the rest of the weights are scaled by 1/(1-p) during training. The use of ensembles of models to improve uncertainty estimates, rather than just improving overall accuracy, has been shown to be effective in the computer vision task of image classification Lakshminarayanan et al. (2017), hence we follow that approach.

The probability associated with a prediction of the ensemble of the 10 models is aggregated as follows:

$$p(y = c) = \frac{1}{M} \Sigma_{i=1}^{M} p(y_{\theta_i} = c | x, \theta_i)$$

The predicted class of the ensemble is

$$\arg\max_{c} p(y = c) = \frac{1}{M} \Sigma_{i=1}^{M} p(y_{\theta_i} = c | x, \theta_i)$$

Where M is the number of models in the ensemble ; $\theta_i$ is the $i^{th}$ model; $y_{\theta_i}$ is the output predicted by the $i^{th}$ model; x is the input; c is the predicted class.

### 2.4 Large scale PPI function extraction from PubMed abstracts

We extracted the complete collection of ~18 million abstracts available in PubMed as of April 2019, using FTP (ftp://ftp.ncbi.nlm.nih.gov//pubmed/baseline/). We applied GNormPlus (Wei 2015) to recognize the proteins and normalize them to UniProt identifiers. We then apply the PPI-BioBERT x10 model to extract PPI functions from the entire PubMed abstracts. In order to extract high quality PPI functions, we use thresholds determined separately for each interaction type, as shown in Table 3. We choose higher thresholds for interactions types that have produced larger number of PPIs to increase precision and lower thresholds for interaction types with low recall. We sample a subset of predicted typed PPIs for human verification to manually assess quality.

## 3 Results

### 3.1 Dataset

The final weakly supervised dataset with the fields [Participant1-Uniprot, Participant2-Uniprot, PubMed abstract, Normalized Abstract, Class]. The class labels are acetylation, methylation, demethylation, phosphorylation, dephosphorylation, ubiquitination deubiquitination, and negative (all negative samples combined). The use of normalized abstracts, where the gene names are replaced by their corresponding UNIPROT identifiers, is necessary to quickly identify the protein pair that is evaluated for a machine learning model and to make the output compatible with various types of downstream protein-protein network analysis. This means that recognition of protein names and normalization to database identifiers is a crucial step in extracting typed interactions between a pair of proteins.

The distribution of interaction types and positive/negative samples as shown in Table 1, with phosphorylation forming almost 74% of the positive samples in the training set. The positive sample rate in the training set is 35 % and the overall positive rate is 34%. Interaction types such as such as demethylation only have 6 positive samples in total across training, test and validation. Similarly, deubiquitination and ubiquitination have 15 and 13 positive samples respectively.

**Table 1 - Train/Test/ Val Positive and negative samples for each interaction type**

| | **Train** | | **Val** | | **Test** | | **Total** | |
|---|---|---|---|---|---|---|---|---|
| | - | + | - | + | - | + | - | + |
| acetylation | 79 | 29 | 14 | 4 | 27 | 7 | 120 | 40 |
| methylation | 72 | 48 | 3 | 9 | 16 | 7 | 91 | 64 |
| demethylation | 6 | 4 | 0 | 0 | 10 | 2 | 16 | 6 |
| phosphorylation | 1814 | 629 | 257 | 79 | 497 | 165 | 2568 | 873 |
| dephosphorylation | 362 | 108 | 32 | 11 | 101 | 20 | 495 | 139 |
| ubiquitination | 34 | 9 | 0 | 1 | 25 | 3 | 59 | 13 |
| deubiquitination | 17 | 12 | 10 | 1 | 12 | 2 | 39 | 15 |
| **Total** | 2384 | 839 | 316 | 105 | 688 | 206 | 3388 | 1150 |

### 3.2 Results on the PPI with function dataset

The ensemble BioBERT model (PPI-BioBERT x10) substantially outperforms a single PPI-BioBERT model on the typed PPI extraction task, achieving a near-19 point absolute improvement in F1 score (54.0 vs 35.4). This performance improvement using ensembles is seen in the untyped PPI dataset as well, as shown in Table 2. The untyped version of the PPI typed dataset, which only requires binary classification, obtains an F-score of 71.7 using the ensemble model.

**Table 2 - PPI dataset scores. (P=Precision, R=Recall)**

| Dataset | Model | P | R | F1 |
|---|---|---|---|---|
| Untyped PPI | NoType-PPI-BioBERT | 74.0 (5.3) | 65.0 (6.0) | [2]69.3 (3.7) |
| Untyped PPI | NoType-PPI-BioBERTx10 | 80.6 | 64.6 | **71.7** |
| Typed PPI | PPI-BioBERT | 34.4 (6.2) | 38.3 (6.9) | [1]35.4 (5.4) |
| Typed PPI | PPI-BioBERTx10 | 66.4 | 51.4 | **54.0** |

[1.] For the typed PPI multi-class dataset, we report the F1-score macro average on test set. Micro averaged F1-score is 85.6 on the test set. The run selected is based on the best performance on the validation set which has a F1-macro score of 57.3, p-macro= 56.7 and r-macro =62.1, F1-micro=81.2

[2] This is the F-score binary on the test set. The corresponding validation set scores (P, R, F1) are 73.4, 76.2, 74.8

### 3.3 Results - Large scale PPI extraction from PubMed abstracts

We extracted 439,647 PPI in total from PubMed abstracts, prior to applying interaction-specific thresholds to select from the predictions of the PPI-BioBERT x10 ensemble model. In order to maintain a high quality, applying interaction-specific thresholds to balance between precision and recall of the extraction of a given interaction, we retain 3253 PPI interactions as shown below in Table 3. We were not able to extract any instances of demethylation, most likely due to the fact we had only 4 training samples.

**Table 3 - New interactions extracted from entire PubMed abstracts using interaction type specific thresholds**

| Interaction type | # PPI | Cut-off | # Post PPI | # Sample | % human precision |
|---|---|---|---|---|---|
| acetylation | 2835 | 0.83 | 14 | 14 | 87.5 |
| methylation | 13143 | 0.85 | 1888 | 17 | 17.6 |
| demethylation | 1 | 0.0 | 0 | 0 | - |
| phosphorylate | 404850 | 0.98 | 1099 | 17 | 23.5 |
| dephosphoryl. | 17319 | 0.85 | 133 | 10 | 60.0 |
| ubiquitinat. | 354 | 0.3 | 31 | 10 | 50.0 |
| deubiquitin. | 1145 | 0.5 | 88 | 10 | 50.0 |
| **Total #** | **439647** | **-** | **3253** | **78** | **46** |

The human verified precision (see supplementary Predictions.xlsx) within a randomly selected subset of 78 PPIs from these predictions for each interaction type is shown in Table 3, with an overall precision of 46%. Acetylation has the highest precision of 87.5%. Methylation and phosphorylation have the lowest precision. We find that the main source of error for methylation prediction is from abstracts that describe DNA methylation at a gene-specific genomic locus rather than describing methylation of a protein (the PTM of interest). We also think that the high error rate in phosphorylation despite having the highest number of training samples could be related to noisy training data as discussed in Section 4.1. This level of false positive rate for phosphorylation is not obvious in our test set, as it requires manual verification, where only ~15% of the phosphorylation predictions (17 out of 110 see confusion matrix in supplementary table S1), arise from negative classes being predicted as phosphorylation. In 2.5% of the sampled cases the human curator could not unambiguously verify the correctness of the predicted PPI by reading the abstract alone.

# 4 Discussion

.

## 4.1 Noisy labels due to weak supervision

When analyzing the predictions in the test set, we found two main sources of error related to noisy, weakly supervised data. The first source is incomplete annotation, where not all relationships between the proteins in the abstract are correctly annotated based on our weak supervision methodology. For instance, the IntAct database indicates that there is a "direct interaction" relationship between protein OXSR1_HUMAN (O95747) and PAK1 (Q13153) described in PMID 14707132. Our normalized PubMed abstract contains the phrase "*O95747 phosphorylated threonine 84 in the N-terminal regulatory domain of Q13153*" and the model correctly identifies the relationship between O95747 & Q13153 as "phosphorylation". This is counted as a False Positive, although it is clearly correct.

The second source of noise is where the relationship is not clearly described in the abstract between proteins but is annotated in IntAct; in this case the interaction may be described in the body of the full text article. For example, phosphorylation between Q13315 (ataxia-telangiectasia mutated) and Q12888 (53BP1) is annotated with reference to PMID 22621922 but it is not clearly described in the abstract. This case would be treated as a positive example, but may be a difficult relationship for the model to identify if the interaction itself is not described in the abstract.

## 4.2 Large scale extraction

A technical limitation is the effectiveness of the underlying named entity recognizer and gene name normalizer, in our case GNormPlus (Wei 2015). To utilize the information in IntAct for weak supervision, we must be able to connect the information captured in the database directly to the texts. This assumes accurate recognition of protein mentions in the text and normalization to the relevant protein identifier for the appropriate biological organism. If the protein name is not identified or associated with the incorrect protein identifier by the GNormPlus tool, then the quality of PPIs annotated on the basis of the database would be negatively affected.

During large scale PPI extraction, the trade-off between quantity (recall) and quality (precision) is an aspect that requires careful consideration. If interaction-specific probability thresholding is not applied, then ~400,000 PPIs can be extracted as shown in Table 3. However, the quality of these extractions is affected by a) weakly supervised training data (noisy labels) and b) the reliability of predictions of machine learning models in general. Probability thresholding is one way to improve the quality; the higher the threshold the more likely that the predictions are of high quality. However, the ability to rely on the confidence score, even with gold standard labels, associated with a prediction of a neural network itself is an active area of research. The use of ensemble models results in substantially increased F-score of 54.0 (PPI-BioBERTx10, Table 2), where basing the confidence score on the average score of models can improve uncertainty estimation (Lakshminarayanan 2017). Noisy labels introduce another level of complexity for machine learning; unless the results are sampled through a human annotator it is difficult to gauge the true performance of the model.

# 5 Conclusions

We created a weakly supervised training dataset for extracting 7 types of PPI function – phosphorylation, dephosphorylation, methylation, demethylation, ubiquitination, deubiquitination, and acetylation – from PubMed abstracts by leveraging IntAct database. We used this dataset to train an ensemble model, PPI-BioBERT-x10, to improve uncertainty estimation of our deep learning model. We applied our model to text-mine 18 million PubMed abstracts, extracting 3253 high confidence PPI interaction functions using the PPI-BioBERT-x10 ensemble we trained. Human verification of randomly selected subset of extracted PPIs had a precision of 46%.

Our work on typed PPI function extraction from PubMed abstracts demonstrates that uncertainty estimation is crucial for maintaining the quality of PPI extraction during large scale application, and needs to be explored further especially in the context of noisy labels. Nonetheless, in the 3253 high confidence typed interactions we extract, we increased (by 35%) the number of acetylation interactions available for future training, and demonstrate the efficacy of this approach in targeted abstract curation to expand database annotation of function.

# 6 Competing interests

We certify that there are no financial and non-financial competing interests regarding the materials discussed in the manuscript.

# 7 Acknowledgements

We would like to thank the authors of Hsieh (2017) for sharing their code and preprocessed dataset to assist us in verifying their experiment. This work was supported by the Australian Research Council [DP190101350 to K.V.].